\documentclass[11pt]{article}
\usepackage{acl}
\usepackage{times}
\usepackage{latexsym}
\usepackage[T1]{fontenc}
\usepackage[utf8]{inputenc}
\usepackage{microtype}
\usepackage{inconsolata}
\usepackage{amsmath}
\usepackage{amssymb}
\usepackage{amsfonts}
\usepackage{booktabs}
\usepackage{graphicx}

\title{Decoding Transformer Dynamics via Koopman Subspaces}

\author{Subhradeep Kayal \quad Ali Dashti \\
  Amazon\thanks{This work was done outside of Amazon.} \\
  \texttt{deep.kayal@pm.me}}

\begin{document}
\maketitle

\begin{abstract}
We apply Koopman operator theory and Dynamic Mode Decomposition (DMD) to characterise how large language models (LLMs) process different input topics. Treating the layer-to-layer evolution of hidden states as a dynamical system, we fit a linear Koopman operator and take the subspace spanned by its dominant modes as a per-topic \textbf{spectral fingerprint}. Four findings emerge. First, the depth map is \emph{near-linear}: a single linear operator predicts each next-layer state to within $\approx 18\%$ relative error (interior variance explained $\approx 0.99$). Second, over the 57 MMLU subjects the fingerprint is \emph{topic-discriminative}, matching the MMLU taxonomy at $0.72$ nearest-neighbour purity and each subject's semantic neighbourhood at $0.75$ any-hit@3. Third, it is a \emph{reproducible} signature (split-half self-retrieval $\approx 0.97$). Finally, these findings are consistent across seven base models (Qwen2.5 0.5-7B, Llama-3 1-8B), with a suggestive trend of alignment rising with MMLU accuracy. The spectral fingerprint thus provides a probe of how transformers process different topics across depth. Code is available at \url{https://github.com/DeepK/koopman-llm-eacl27}.
\end{abstract}

\section{Introduction}

Large language models process all inputs through the same fixed sequence of transformer layers, regardless of input complexity or domain. While prior work has studied \textit{how much} and how fast each layer changes a representation (via cosine similarity between consecutive hidden states \cite{men2024shortgpt}, angular distance \cite{gromov2024unreasonable}, or velocity and acceleration profiles \cite{fernando2025transformer}), these measures do not capture \textit{how} the representation is transformed: which directions it moves along, and whether features are amplified, suppressed, or mixed together. They therefore cannot tell whether two inputs are transformed in the same way or only by the same amount.

We propose a different lens: the hidden states a prompt produces across layers form the trajectory of a \textbf{dynamical system}, and \textbf{Koopman operator theory} gives its spectral structure. The Koopman operator linearises nonlinear dynamics by acting on observable functions, and its eigenvalues describe the fundamental modes of the system. Applied to an LLM, the corresponding \textbf{modes} are feature directions in hidden-state space along which representations are transformed across depth.

Our central object is the \textbf{spectral fingerprint}: the subspace spanned by a topic's dominant Koopman modes, an embedding-like signature of how the model processes that topic. We obtain it from \emph{linear} DMD, which we show captures the one-step layer-to-layer dynamics well, suggesting that the bulk of the LLM's layers are nearly linear. Comparing these mode subspaces across topics defines a \textbf{dynamics space} in which two topics are ``close'' if the model transforms them along similar directions, regardless of whether they are semantically related. This lets us ask questions that magnitude measures cannot: which topics a model processes alike, including some that are not semantically close (Section~\ref{sec:taxonomy}); whether a topic has a stable processing signature at all (Section~\ref{sec:reliability}); and where in depth a linear description holds and where it breaks down (Section~\ref{sec:recon57}).

\paragraph{Contributions.}
\begin{itemize}
    \item We introduce the \textbf{mode-subspace spectral fingerprint}, a per-topic representation built from a Koopman operator fitted along depth, together with a Grassmann distance for comparing fingerprints across topics.
    \item The transformer's depth map is \textbf{near-linear}: a single linear operator captures each layer transition (interior $R^2 \approx 0.99$) and outperforms a nonlinear kernel lift, with nonlinearity concentrated at the input and output boundaries, so its Koopman spectrum is a well-grounded description of the one-step dynamics.
    \item On Qwen2.5-3B over 57 MMLU subjects the fingerprint carries topic identity: its neighbourhoods are category-aligned with the MMLU taxonomy ($0.72$ 1-NN purity) and overlap semantic neighbourhoods at nearly $5\times$ chance, and where it departs from semantics it groups subjects by computation (quantitative subjects clustering around machine learning).
    \item We show the fingerprint is a \textbf{reproducible} signature (split-half self-retrieval $0.97$) that is \textbf{consistent} across seven base models spanning Qwen and Llama and $0.5$-$8$B in scale, with a suggestive upward trend in alignment with the models' MMLU accuracy.
\end{itemize}

\section{Related Work}
\label{sec:related_work}

\subsection{Transformers as Depth-Wise Dynamical Systems}
Viewing deep networks as discretised ordinary differential equations (ODEs) has a rich foundation~\cite{weinan2017proposal, chen2018neural}; for Transformers, this lens treats the residual stream's layer-wise updates as a trajectory in representation space. Recently, \citet{fernando2025transformer} formalise the residual stream as a spatial dynamical system, reporting systematic velocity and acceleration profiles across depth. These measures describe how much and how fast representations change; we instead fit an explicit linear operator to each topic's transitions and use its dominant modes to identify the directions of change. A complementary line finds consecutive layers near-linearly related and partly redundant: adjacent representations align almost perfectly under a linear map \citep{razzhigaev2024linear}, and many layers can be removed with little loss \citep{gromov2024unreasonable, men2024shortgpt}, which has motivated architectures that allocate depth conditionally per token \citep{elhoushi2024layerskip, raposo2024mixture}. These results motivate treating each layer-to-layer step as approximately linear, which our fingerprint exploits.

\subsection{Dynamic Mode Decomposition and Koopman Theory in language models}
DMD and Koopman theory are increasingly used to study foundation models, though along dimensions orthogonal to ours. One line models physical or fluid dynamics with attention-based architectures~\cite{learningkoopman2026}. Closer to text, \citet{akrout2023representations} applies DMD to extract semantic modes from embeddings across sequence length, detecting structural drift and hallucination, and \citet{kayal2019eigensent} apply higher-order DMD across word sequences to build spectral sentence embeddings. Most related, \citet{kim2026rksp} profile the Koopman spectrum of a transformer's residual dynamics at initialisation to predict training instability, treating depth-wise evolution as a dynamical system whose spectrum is diagnostic.

The first two works apply spectral analysis along the \textit{sequence dimension} (token-to-token or sentence-to-sentence). Like \citet{kim2026rksp}, we apply DMD along the \textit{depth dimension} (layer-to-layer) within a single forward pass; they fit a separate operator per layer as a training-stability diagnostic, whereas we fit one operator across all layers of a trained model, per input topic, so that topics can be compared in a common dynamics space.

\subsection{Depth-Wise Spectral Analysis and Model Compression}
Depth-wise DMD has emerged in computer vision for architectural efficiency: recent work applies it to Vision Transformers (ViTs) to test whether sequential non-linear blocks can be approximated by a low-rank linear operator for block pruning and compression~\cite{vitdepth2026}. We use the same depth-wise decomposition for a different purpose. That work fits the operator to replace blocks of layers and so compress the model; we fit it per topic and compare the resulting fingerprints.

\subsection{Mechanistic and Structural Interpretability}
Our fingerprints relate to work mapping how internal representations evolve across layers. Structural probes such as the Logit Lens~\cite{nostalgebraist2020} and Tuned Lens~\cite{belrose2023eliciting} decode hidden states by projecting them into vocabulary logit space, and representation engineering~\cite{zou2023representation} isolates directional vectors for high-level concepts. Whereas these identify \textit{what} a layer contains, typically per concept, our approach models how representations are transformed between layers and summarises this for a whole topic.

\section{Methodology}

\subsection{Koopman Operator Theory}
\label{sec:koopman}

Consider a nonlinear dynamical system $\mathbf{x}_{t+1} = \mathbf{F}(\mathbf{x}_t)$ where $\mathbf{x}_t \in \mathbb{R}^D$. The \textbf{Koopman operator} $\mathcal{K}$~\citep{koopman1931hamiltonian} is an infinite-dimensional linear operator that acts on \textit{observable functions} $g: \mathbb{R}^D \to \mathbb{C}$:
\begin{equation}
    (\mathcal{K} g)(\mathbf{x}) = g(\mathbf{F}(\mathbf{x}))
\end{equation}

$\mathcal{K}$ is \textit{linear} even when $\mathbf{F}$ is nonlinear; linearity is achieved by lifting into the (infinite-dimensional) space of observables. Its spectral decomposition underpins modern operator-theoretic analysis of dynamical systems \citep{mezic2005spectral, brunton2022modern}. The Koopman eigenvalues $\lambda_j$ and eigenfunctions $\varphi_j$ satisfy:
\begin{equation}
    \mathcal{K} \varphi_j = \lambda_j \varphi_j
\end{equation}

We interpret the transformer's layer-to-layer map as $\mathbf{F}$, with hidden states $\mathbf{h}_l \in \mathbb{R}^D$ as the state at ``time'' (depth) $l$. Each eigenfunction is paired with a \textbf{Koopman mode} $\boldsymbol{\phi}_j$, an eigenvector in hidden-state space that identifies a coherent feature direction the dynamics act on.

\paragraph{Relation to the layer Jacobian.} A more familiar linearisation of a layer is its Jacobian $\mathbf{J}_l(\mathbf{h}) = \partial \mathbf{F}_l / \partial \mathbf{h}$, the first-order Taylor approximation of layer $l$ around a single hidden state $\mathbf{h}$. It is accurate near that state but changes from input to input, so it describes the layer at one point. The DMD operator (Section~\ref{sec:dmd}) is instead the least-squares best linear map over all of a topic's layer transitions, across every prompt and depth. Its modes are therefore the directions along which the model transforms that topic on average, which makes them a topic-level descriptor where a pointwise Jacobian is not.

\subsection{Dynamic Mode Decomposition (DMD)}
\label{sec:dmd}

DMD \citep{schmid2010dmd} provides a finite-dimensional approximation to the Koopman operator from data \citep{rowley2009spectral}; our formulation follows \emph{exact} DMD \citep{tu2014dmd}. Given snapshot pairs $(\mathbf{X}, \mathbf{Y})$ where $\mathbf{Y} \approx \mathbf{A}\mathbf{X}$:

\paragraph{Data construction.} For a set of $N$ input prompts with hidden states at $T$ layers, we form:
\begin{align}
    \mathbf{X} &= [\mathbf{h}_0^{(1)}, \ldots, \mathbf{h}_{T-2}^{(1)}, \ldots, \mathbf{h}_{T-2}^{(N)}] \in \mathbb{R}^{D \times M} \\
    \mathbf{Y} &= [\mathbf{h}_1^{(1)}, \ldots, \mathbf{h}_{T-1}^{(1)}, \ldots, \mathbf{h}_{T-1}^{(N)}] \in \mathbb{R}^{D \times M}
\end{align}
where $M = N(T-1)$ is the total number of transition samples.

\paragraph{Whitened DMD.} To ensure scale invariance across hidden dimensions, we apply whitening before decomposition. Let $\mathbf{U}_x \boldsymbol{\Sigma}_x \mathbf{V}_x^\top = \text{SVD}(\mathbf{X})$. The whitening transform is:
\begin{equation}
    \mathbf{W} = \boldsymbol{\Sigma}_x^{-1/2} \mathbf{U}_x^\top
\end{equation}
We then compute $\tilde{\mathbf{X}} = \mathbf{W}\mathbf{X}$ and $\tilde{\mathbf{Y}} = \mathbf{W}\mathbf{Y}$. We apply whitening for numerical conditioning; for linear DMD it is a similarity transform of the fitted operator (modes are mapped back to the original space), so it leaves the fingerprint unchanged (Section~\ref{sec:ablation}).

\paragraph{Rank-truncated SVD.} Compute $\tilde{\mathbf{U}}_r \tilde{\boldsymbol{\Sigma}}_r \tilde{\mathbf{V}}_r^\top \approx \tilde{\mathbf{X}}$ (rank-$r$ truncation). The reduced Koopman operator is:
\begin{equation}
    \tilde{\mathbf{A}} = \tilde{\mathbf{U}}_r^\top \tilde{\mathbf{Y}} \tilde{\mathbf{V}}_r \tilde{\boldsymbol{\Sigma}}_r^{-1} \in \mathbb{R}^{r \times r}
\end{equation}

\paragraph{Eigendecomposition.} Solving $\tilde{\mathbf{A}} \mathbf{w}_j = \lambda_j \mathbf{w}_j$ yields DMD eigenvalues $\{\lambda_j\}_{j=1}^r$ and modes $\boldsymbol{\Phi} = \tilde{\mathbf{Y}} \tilde{\mathbf{V}}_r \tilde{\boldsymbol{\Sigma}}_r^{-1} \mathbf{W}_e$, mapped back to original space via $\mathbf{W}^{-1}$.

\subsection{Spectral Fingerprinting}

For a topic $t$ with prompts $\{p_i\}_{i=1}^N$, we extract the full-depth trajectory and apply linear DMD (Section~\ref{sec:ablation}) to obtain eigenvalues $\{\lambda_j\}$ and modes $\{\boldsymbol{\phi}_j\}$. We project every snapshot onto the modes, $\mathbf{B} = \boldsymbol{\Phi}^{\dagger}\mathbf{X} \in \mathbb{C}^{r \times M}$, so that row $\boldsymbol{b}_j$ holds mode $j$'s coefficient on each of the $M$ snapshots across all prompts and layers. The energy of mode $j$ is $e_j = \|\boldsymbol{\phi}_j\| \cdot \|\boldsymbol{b}_j\|$, its total contribution to reconstructing the trajectories, rather than its amplitude in the initial state alone.

\paragraph{Fingerprint (mode subspace).} The fingerprint is the subspace spanned by the top-$K$ modes ranked by energy:
\begin{equation}
    \mathcal{S}^{(t)} = \mathrm{span}\big(\boldsymbol{\phi}_{(1)}, \ldots, \boldsymbol{\phi}_{(K)}\big) \subset \mathbb{C}^{D},
\end{equation}
represented by an orthonormal basis $\mathbf{Q}^{(t)} \in \mathbb{C}^{D\times K}$ (from a QR factorisation). This is an embedding-like object (akin to a word vector, but a $K$-dimensional subspace rather than a single direction) and it is invariant to each mode's arbitrary scale, complex phase, and ordering.

\paragraph{Dynamics-space distance.} Similarity between two fingerprints is the Grassmann (principal-angle) distance \citep{bjorck1973angles, edelman1998geometry} between their subspaces:
\begin{equation}
   \begin{split}
       d_{\text{dyn}}(t_1, t_2) &= \Big(\textstyle\sum_{i=1}^{K} \theta_i^2\Big)^{1/2}, \\
       \cos\theta_i &= \sigma_i\!\big(\mathbf{Q}^{(t_1)\dagger}\mathbf{Q}^{(t_2)}\big)
   \end{split}
   \end{equation}
where $\{\sigma_i\}$ are the singular values of the cross-basis matrix and $\{\theta_i\}$ the principal angles. Equivalently one may use the projection cosine $\tfrac{1}{K}\sum_i\cos^2\theta_i \in [0,1]$.

\paragraph{Surface-space distance.} As a semantic reference we embed each subject's label text with a sentence-transformer \citep{reimers2019sentencebert} and take cosine distance, $d_{\text{emb}}(t_1,t_2) = 1 - \cos(\mathbf{e}^{(t_1)}, \mathbf{e}^{(t_2)})$.

\section{Experiments}

\subsection{Setup}

\paragraph{Models.} We evaluate seven base models across two axes (results in Section~\ref{sec:crossmodel}):
\begin{itemize}
    \item \textbf{Scale}: Qwen2.5-0.5B / 1.5B / 3B / 7B
    \item \textbf{Family}: Qwen2.5, Llama-3.2 (1B, 3B), Llama-3.1-8B
\end{itemize}

\paragraph{Data.} We use MMLU \cite{hendrycks2021mmlu} with its 57 subjects spanning STEM, humanities, social sciences, and other categories. We extract 100 prompts per subject from the test split: we cap at 100 because the smallest subject's test set contains only 100 examples, so 100 is the largest sample size available uniformly across all 57 subjects.

\paragraph{Compute.} All experiments run on a single CPU machine (32-core Intel Xeon, 64\,GB RAM).

\subsection{Research Questions}

We organise the study around the following research questions.
\begin{itemize}
    \item \textbf{RQ1.} Can a Koopman/DMD operator capture the layer-to-layer dynamics of a transformer?
    \begin{itemize}
        \item \textbf{RQ1a.} Is a linear operator sufficient?
    \end{itemize}
    \item \textbf{RQ2.} Does the mode-subspace fingerprint encode meaningful semantic structure?
    \begin{itemize}
        \item \textbf{RQ2a.} How many modes does a stable fingerprint need?
        \item \textbf{RQ2b.} Is the resulting fingerprint a reliable, reproducible signature?
    \end{itemize}
    \item \textbf{RQ3.} Is the fingerprint's quality consistent across model families and scales? That is, is a reproducible, category- and semantically-aligned fingerprint present in every model tested, or only in a particular family or size?
\end{itemize}

\section{Results and Analysis}
\label{sec:prelim}

\subsection{How Well Does a Koopman Operator Capture the Dynamics? (RQ1)}
\label{sec:recon57}

We quantify how well the fitted operator captures the layer-to-layer dynamics with a \textbf{mean per-step prediction error}: each layer transition is predicted from the \emph{true} previous state and the relative $L_2$ error is averaged over depth, so errors do not compound. This isolates the \emph{one-step} dynamics, the property relevant to a meaningful eigenvalue spectrum. Because the operator is fitted jointly to every transition of every prompt, a low error means that one shared linear map fits the whole distribution of a topic's layer transitions. This is a stronger statement than linearity near a single state (the Jacobian view of Section~\ref{sec:koopman}). On \textbf{Qwen2.5-3B} (36 layers, $D{=}2048$, 100 prompts per subject), linear DMD (rank $100$) applied to all $57$ MMLU subjects attains a mean per-step error of $0.184$ (range $0.121$-$0.207$, median $0.184$), a per-step variance explained ($R^2 = 1 - \text{err}^2$) of $0.91$ averaged over layers (range $0.87$-$0.94$).

\paragraph{Interpreting the error.} A per-step error of $0.184$ means that, given the true hidden state at a layer, the operator predicts the next layer's state to within $\approx 18\%$ relative $L_2$ error on average. Because consecutive residual-stream states are similar ($\mathbf{h}_{l+1} = \mathbf{h}_l + \delta$), part of this accuracy reflects representational \emph{continuity}; we quantify this with an identity baseline that predicts $\hat{\mathbf{h}}_{l+1} = \mathbf{h}_l$. On the five subjects used for the decomposer comparison (Section~\ref{sec:ablation}), where linear DMD's error is $0.190$ (vs.\ $0.184$ over all 57), the identity baseline yields $0.245$; linear DMD improves on it by $\approx 22\%$ (consistent across topics, $20$-$25\%$), which is transformation beyond continuity.

\paragraph{Depth structure.} Table~\ref{tab:perstep} breaks the error down by layer, averaged over subjects.

\begin{table}[!htbp]
\centering
\small
\begin{tabular}{@{}lc@{}}
\toprule
\textbf{Depth transition} & \textbf{Variance explained} \\
\midrule
$\mathbf{h}_0 \!\to\! \mathbf{h}_1$ (embedding ingest) & $-0.04$ \\
$\mathbf{h}_1 \!\to\! \mathbf{h}_2$ & $0.35$ \\
$\mathbf{h}_2 \!\to\! \mathbf{h}_3$ & $0.48$ \\
\textbf{interior ($\mathbf{h}_3 \!\to\! \mathbf{h}_{33}$)} & $\boldsymbol{\approx 0.99}$ \\
$\mathbf{h}_{35} \!\to\! \mathbf{h}_{36}$ (final layer) & $0.73$ \\
\midrule
interior mean (steps $1$-$34$) & $0.94$ \\
\bottomrule
\end{tabular}
\caption{Per-step variance explained by linear DMD on Qwen2.5-3B, averaged over all 57 subjects.}
\label{tab:perstep}
\end{table}

\begin{figure}[t]
\centering
\includegraphics[width=1.0\columnwidth]{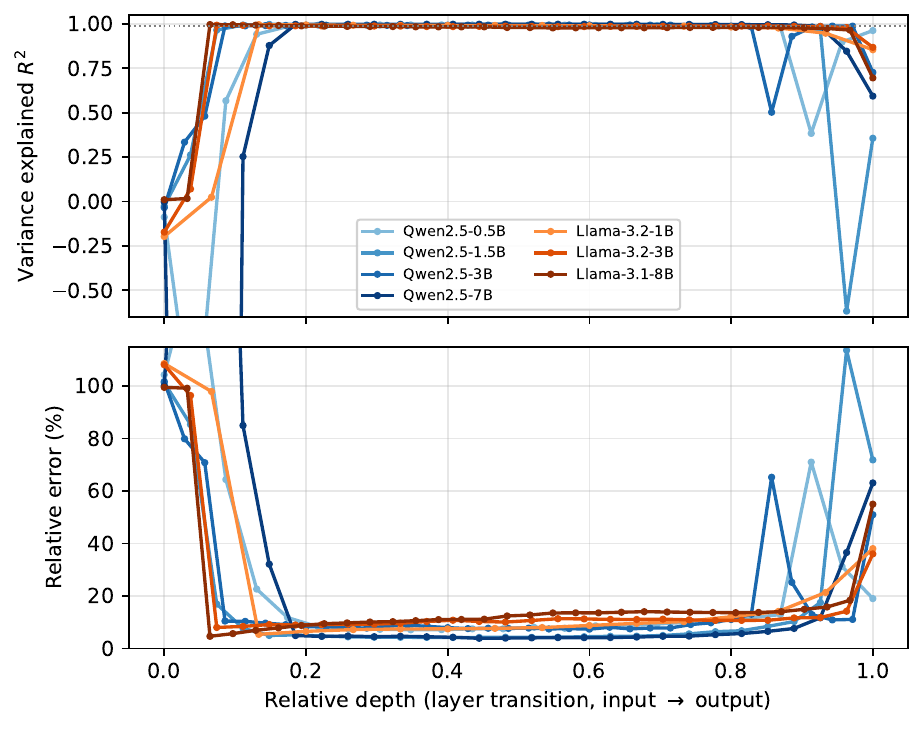}
\caption{\textbf{The layer-to-layer map is near-linear in the interior across all models.} Per-step variance explained $R^2$ (top) and relative prediction error (bottom) of the linear Koopman operator, averaged over the 57 MMLU subjects, against relative depth (input transition at $0$, output at $1$) for the seven base models (Qwen2.5 in blue, Llama in orange). The interior is flat at $R^2 \approx 0.99$ ($\approx$5--15\% error) for every model, while both boundaries (embedding ingestion and final-layer projection) are strongly nonlinear.}
\label{fig:near-linearity}
\end{figure}

The first transition ($\mathbf{h}_0 \to \mathbf{h}_1$) has variance explained near zero (slightly negative): mapping raw token embeddings into the residual stream is a large, strongly nonlinear ``ingestion'' step that a single linear operator fit to the whole trajectory cannot capture. The final transition ($\mathbf{h}_{35} \to \mathbf{h}_{36}$) is also markedly less linear ($R^2 \approx 0.73$), consistent with output-oriented reshaping in the last layer. Between these, layers $3$-$33$ are captured at $R^2 \approx 0.99$: \textbf{the transformer's depth map is near-linear in its bulk and concentrates its nonlinearity at the input and output boundaries}, a structural property the fingerprint summarises. This interior near-linearity is independently corroborated by \citet{razzhigaev2024linear}, who find near-perfect linear alignment (Procrustes similarity $\approx 0.99$) between adjacent decoder-layer representations, strongest in the middle layers. Figure~\ref{fig:near-linearity} shows the same profile for all seven models (Section~\ref{sec:crossmodel}).

\subsection{Is a Linear Operator Sufficient? (RQ1a)}
\label{sec:ablation}

Since each transformer layer is nonlinear, one might expect a single linear operator to be inadequate and a nonlinear lift to help. We therefore compare linear DMD against \textbf{kernel DMD} \citep{williams2015edmd, williams2015kernel}, which lifts the data into a reproducing-kernel Hilbert space (RKHS) where the dynamics may become more linear. With a Gaussian RBF kernel $k(\mathbf{x},\mathbf{x}')=\exp(-\|\mathbf{x}-\mathbf{x}'\|^2/2\sigma^2)$ and Gram and cross-kernel matrices $\mathbf{G}_{ij}=k(\mathbf{x}_i,\mathbf{x}_j)$ and $\hat{\mathbf{A}}_{ij}=k(\mathbf{x}_i,\mathbf{y}_j)$, the kernel-DMD eigenvalues solve $(\mathbf{G}+\epsilon\mathbf{I})^{-1}\hat{\mathbf{A}}\,\boldsymbol{\alpha}_j=\lambda_j\boldsymbol{\alpha}_j$, with $\epsilon$ regularising the (possibly ill-conditioned) Gram matrix. Using the same per-step prediction error, we sweep the kernel bandwidth $\sigma$ (as a multiple of the median heuristic) and global state whitening, with per-state normalisation disabled throughout, on five randomly chosen MMLU subjects (Table~\ref{tab:ablation}).

\begin{table}[!htbp]
\centering
\small
\begin{tabular}{@{}lcc@{}}
\toprule
\textbf{Configuration} & \textbf{no whiten} & \textbf{whiten} \\
\midrule
\textbf{Linear DMD} & \multicolumn{2}{c}{\textbf{0.190}} \\
\midrule
Kernel, $\sigma{\times}0.1$ & 0.64 & 0.98 \\
Kernel, $\sigma{\times}0.25$ & 0.81 & 0.93 \\
Kernel, $\sigma{\times}0.5$ & 2.92 & 1.13 \\
Kernel, $\sigma{\times}1.0$ & 5.66 & 5.60 \\
Kernel, $\sigma{\times}2.0$ & 11.83 & 4.66 \\
\bottomrule
\end{tabular}
\caption{Mean per-step prediction error on five random subjects, lower is better. Linear whitening is omitted as it is a basis change that leaves the original-space operator, and hence the error, unchanged (0.190 either way). The five-subject linear value ($0.190$) is close to the all-57 mean ($0.184$, Section~\ref{sec:recon57}).}
\label{tab:ablation}
\end{table}

\textbf{Linear DMD outperforms kernel DMD} ($0.190$ vs.\ a best kernel of $0.64$): the RBF lift does not help. We therefore use linear DMD for fingerprinting; it is also the simplest choice.

\subsection{Does the Fingerprint Encode Meaningful Semantic Structure? (RQ2)}
\label{sec:taxonomy}

We test this on \textbf{Qwen2.5-3B} with $K{=}50$ modes (Section~\ref{sec:modedim}). MMLU groups its 57 subjects into four top-level categories (STEM, humanities, social sciences, other) through the official \texttt{categories.py} mapping; Table~\ref{tab:categories} lists them with their subject counts. We use this grouping as a coarse semantic ground truth: if the fingerprint is meaningful, subjects in the same category should have more similar fingerprints than subjects in different ones.

\begin{table}[!htbp]
\centering
\small
\begin{tabular}{@{}l p{0.58\columnwidth} c@{}}
\toprule
\textbf{Category} & \textbf{Subcategories} & \textbf{\#} \\
\midrule
STEM & physics, chemistry, biology, comp.\ sci., math, engineering & 18 \\
Humanities & history, philosophy, law & 13 \\
Social sciences & politics, culture, economics, geography, psychology & 12 \\
Other & health, business, miscellaneous & 14 \\
\bottomrule
\end{tabular}
\caption{The 57 MMLU subjects grouped into four top-level categories (official \texttt{categories.py} mapping), with the number of subjects in each. Chance 1-NN category purity is $\approx 0.24$.}
\label{tab:categories}
\end{table}

\begin{figure*}[t]
\centering
\includegraphics[width=1.0\textwidth]{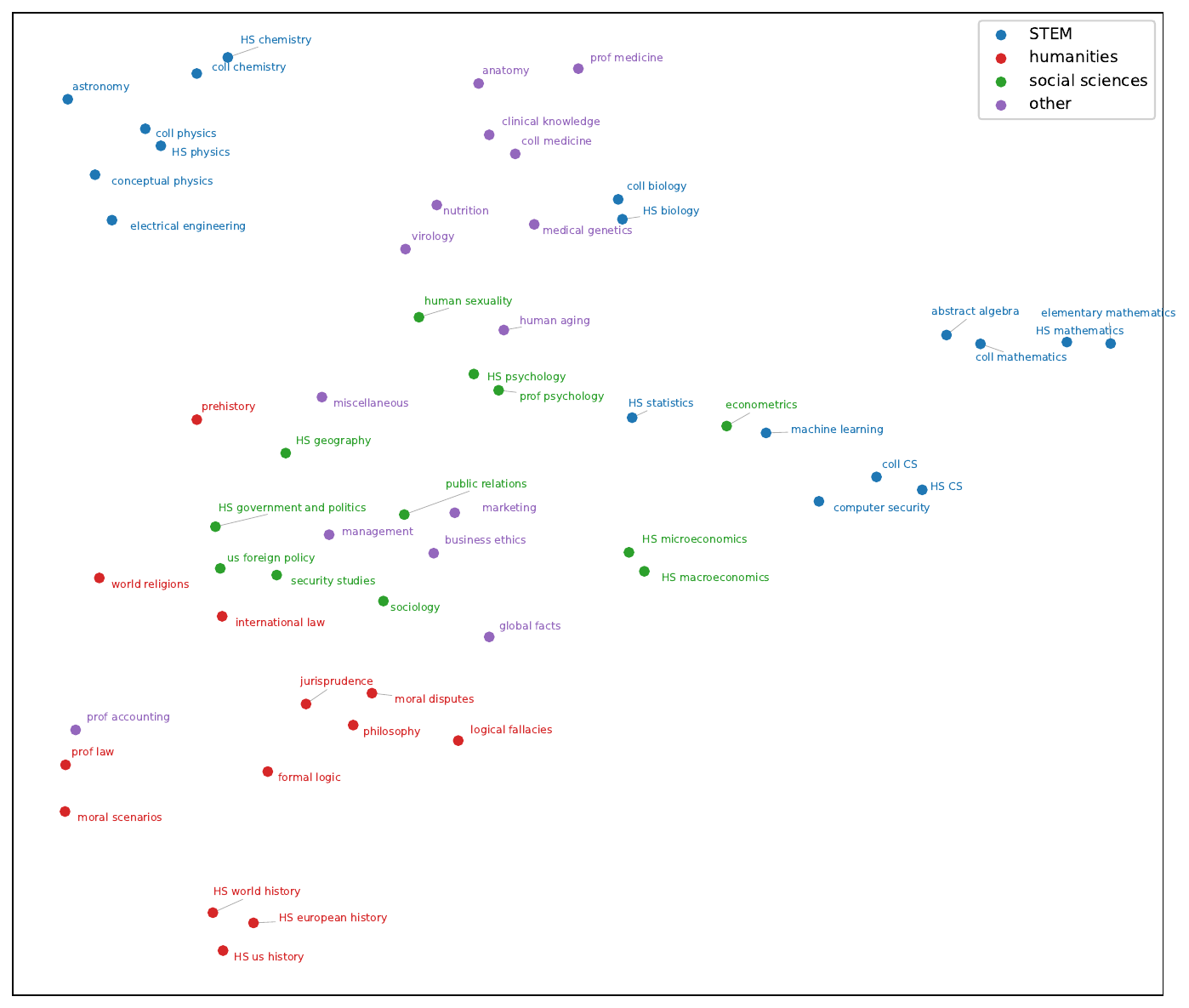}
\caption{\textbf{Mode-subspace fingerprints of the 57 MMLU subjects.} Two-dimensional t-SNE of the $K{=}50$ fingerprint (Grassmann) distances among the 57 subjects for Qwen2.5-3B, with every subject labelled and coloured by top-level category. Same-category subjects largely group together (history at the bottom, physics and chemistry upper-left, medicine and biology upper-centre, pure mathematics at the right), but the salient structure is \emph{cross-category}: the quantitative/algorithmic subjects (machine learning, computer science, computer security, econometrics, statistics, and the economics subjects) form a single cluster, and electrical engineering joins the physics group.}
\label{fig:fingerprint-map}
\end{figure*}

\paragraph{Category alignment.} We quantify category coherence with \textbf{1-NN purity}. For each subject we compute the Grassmann distance between its top-$50$ mode subspace and that of every other subject, identify its single nearest neighbour, and check whether that neighbour belongs to the same category; the 1-NN purity is the fraction of subjects for which it does (chance $\approx 0.24$, given the category sizes). The fingerprint reaches a 1-NN purity of $\mathbf{0.72}$: a subject's nearest neighbour in dynamics space usually belongs to the same MMLU category. A two-dimensional t-SNE of the fingerprint distances (Figure~\ref{fig:fingerprint-map}) makes this visible: same-category subjects occupy distinct but overlapping regions.

\paragraph{Dynamics space vs.\ semantic space.} Category labels are coarse, so we also compare the fingerprint against a continuous semantic reference: does similarity in \emph{dynamics space} (fingerprints) match similarity in \emph{semantic space}? We embed each subject's label text (underscores replaced by spaces, e.g.\ ``high school european history'') with a sentence-transformer (\texttt{all-MiniLM-L6-v2}; \citealp{reimers2019sentencebert}) and take cosine distance, a semantic representation independent of the LLMs under study. Taking each subject's $k$ nearest semantic neighbours as a reference set, we ask how many of its $k$ nearest \emph{fingerprint} neighbours fall in that set (Table~\ref{tab:semantic}): the top-1 agreement (whether the single nearest fingerprint neighbour is also the nearest semantic neighbour) and the any-hit@3 rate (the fraction of subjects for which at least one of the three nearest fingerprint neighbours is also among the three nearest semantic neighbours).

\begin{table}[!htbp]
\centering
\small
\begin{tabular}{@{}lcc@{}}
\toprule
\textbf{Representation} & \textbf{top-1} & \textbf{any-hit@3} \\
\midrule
Mode subspace (fingerprint) & $0.33$ & $0.75$ \\
\midrule
chance & $0.02$ & $0.16$ \\
\bottomrule
\end{tabular}
\caption{Overlap between dynamics-space (the mode-subspace fingerprint) and semantic (label-embedding) neighbourhoods over 57 subjects. ``top-1'' is the fraction sharing the single nearest neighbour; ``any-hit@3'' is the fraction of subjects with $\geq 1$ of their top-3 dynamics neighbours among their top-3 semantic neighbours. Chance rows give the expected overlap under random labelling.}
\label{tab:semantic}
\end{table}

The two spaces are aligned well above chance at both granularities: top-1 agreement is $0.33$, about $16\times$ the $0.02$ chance rate, and any-hit@3 is $0.75$, nearly $5\times$ the $0.16$ chance rate. Table~\ref{tab:nn-examples} shows where the two spaces agree and where they differ. Partial agreement is the expected outcome: there is no single ground truth for semantic similarity, and a label embedding is only one proxy for it. This is why we test against two independent references, the MMLU taxonomy and label embeddings. The alignment also holds over the full distance matrices: across all 57 subjects, fingerprint and semantic distances are positively correlated (Mantel test, \citealp{mantel1967detection}; Spearman correlation $0.20$, $p = 0.0002$, 5000 permutations), so it is not confined to selected subjects.

\begin{table*}[!htbp]
\small
\centering
\begin{tabular}{@{}lll@{}}
\toprule
\textbf{Subject} & \textbf{Semantic top-3} & \textbf{Dynamic top-3 (mode subspace)} \\
\midrule
\multicolumn{3}{@{}l}{\emph{Agreement}} \\
college biology & HS biology, coll.\ medicine, coll.\ chemistry & HS biology, coll.\ medicine, medical genetics \\
college chemistry & HS chemistry, coll.\ physics, coll.\ biology & HS chemistry, coll.\ physics, HS physics \\
clinical knowledge & prof.\ medicine, coll.\ medicine, anatomy & coll.\ medicine, anatomy, nutrition \\
formal logic & logical fallacies, jurisprudence, concept.\ physics & jurisprudence, moral disputes, philosophy \\
\midrule
\multicolumn{3}{@{}l}{\emph{Divergence}} \\
college CS & HS CS, coll.\ math, coll.\ physics & HS CS, computer security, \textbf{machine learning} \\
econometrics & HS macroecon., HS microecon., HS statistics & \textbf{machine learning}, HS statistics, HS microecon. \\
computer security & security studies, coll.\ CS, HS CS & coll.\ CS, \textbf{machine learning}, public relations \\
HS geography & HS world hist., HS math, HS US hist. & security studies, public relations, sociology \\
\bottomrule
\end{tabular}
\caption{Top-3 nearest neighbours of each subject in semantic vs.\ mode-subspace dynamics space ($K{=}50$, Qwen2.5-3B). Agreement cases are tight same-family clusters; among the divergences, quantitative/algorithmic subjects have \emph{machine learning} among their nearest dynamics neighbours. ``HS''~=~high school, ``coll.''~=~college, ``CS''~=~computer science.}
\label{tab:nn-examples}
\end{table*}

\paragraph{Interpretation.} The two spaces \emph{agree} where a subject sits in a tight, hierarchical family (biology, chemistry, medicine, logic), which the model both represents and \emph{processes} similarly. Among the divergences, one pattern stands out: semantically distinct quantitative/algorithmic subjects (\emph{college computer science}, \emph{econometrics}, \emph{computer security}) all have \emph{machine learning} among their nearest dynamics neighbours, and \emph{machine learning}'s own nearest dynamics neighbours are exactly these subjects. One reading is that the model applies a shared quantitative-reasoning computation to them regardless of domain. \emph{High school geography} similarly groups with social and political subjects (\emph{security studies}, \emph{public relations}, \emph{sociology}) rather than the history subjects it is closest to semantically. The fingerprint thus tracks surface semantics closely but, where it departs, groups subjects by \emph{how the model computes over them} rather than by lexical similarity.

\subsection{How Many Modes Does the Fingerprint Need? (RQ2a)}
\label{sec:modedim}

The fingerprint is a $K$-dimensional subspace; how large must $K$ be? Table~\ref{tab:modedim} sweeps $K$ from $1$ (a single mode) to $100$ (the full operator rank) for Qwen2.5-3B, reporting category 1-NN purity and the semantic any-hit@3 (fraction of subjects with $\geq 1$ shared top-3 neighbour).

\begin{table}[!htbp]
\centering
\small
\begin{tabular}{@{}lcc@{}}
\toprule
\textbf{$K$ (modes)} & \textbf{cat.\ 1-NN} & \textbf{any-hit@3} \\
\midrule
$1$ & $0.44$ & $0.19$ \\
$2$ & $0.42$ & $0.21$ \\
$3$ & $0.49$ & $0.35$ \\
$5$ & $0.51$ & $0.32$ \\
$10$ & $0.65$ & $0.61$ \\
$20$ & $0.61$ & $0.65$ \\
$30$ & $0.72$ & $0.74$ \\
$\boldsymbol{50}$ & $\boldsymbol{0.72}$ & $\boldsymbol{0.75}$ \\
$70$ & $0.68$ & $0.74$ \\
$100$ & $0.72$ & $0.74$ \\
\bottomrule
\end{tabular}
\caption{Effect of subspace dimension $K$ on Qwen2.5-3B. Category 1-NN purity (chance $\approx 0.24$) and semantic any-hit@3 (fraction of subjects with $\geq 1$ shared top-3 neighbour; chance $\approx 0.16$). Both rise sharply up to $K{\approx}10$; any-hit@3 peaks around $K{=}50$ and plateaus.}
\label{tab:modedim}
\end{table}

A \textbf{single mode is insufficient}: at $K=1$ the any-hit@3 rate ($0.19$) is near chance, and the leading conjugate pair ($K=2$) is no better. Discriminative structure emerges at $K\geq 3$: category 1-NN purity rises from $0.44$ to $\approx 0.72$ and any-hit@3 from $0.19$ to $\approx 0.75$, with most of the gain by $K\approx10$ and a plateau from $K=50$. The fingerprint therefore needs to be a \emph{subspace}: topic information is spread over many mode directions, not concentrated in the dominant one. We adopt $K=50$ throughout.

\subsection{Is the Resulting Fingerprint a Reliable, Reproducible Signature? (RQ2b)}
\label{sec:reliability}

A fingerprint is only useful if it reflects a stable property of a topic rather than the particular prompts used to compute it. We test this with a \textbf{split-half} reproducibility check on Qwen2.5-3B. For each subject we randomly partition its $100$ prompts into two disjoint halves and fit an independent fingerprint from each. We then perform \emph{self-retrieval}: for each half-A fingerprint we find its nearest neighbour among all $57$ half-B fingerprints (by Grassmann distance) and ask whether it is the \emph{same} subject. We report the split-half self-retrieval rate (the fraction of subjects recovered as their own nearest neighbour; chance $\approx 1/57 \approx 0.02$) and the mean self-retrieval rank ($1$ = perfect).

On Qwen2.5-3B (top-$50$ subspace), split-half self-retrieval is $0.97$ with a mean self-rank of $1.0$, against a chance level of $0.02$: for almost every subject, the nearest half-B fingerprint to its half-A fingerprint is its own, and in the few misses the correct subject ranks close to the top. A fingerprint estimated from $50$ prompts is therefore already specific to its topic, out of $57$ candidates. Section~\ref{sec:crossmodel} shows the same for all seven models.

\subsection{Is Fingerprint Quality Consistent Across Models? (RQ3)}
\label{sec:crossmodel}

We extend the analysis to \textbf{seven} base models spanning \textbf{scale} (Qwen2.5 0.5/1.5/3/7B) and \textbf{family} (Qwen2.5, Llama-3.1/3.2). Each model is fingerprinted identically (linear DMD, rank $100$, top-$50$ subspace, 100 prompts/subject) and run through the same category, semantic-overlap, and split-half analyses (Sections~\ref{sec:taxonomy} and~\ref{sec:reliability}). Table~\ref{tab:models} reports the metrics with published 5-shot base MMLU.

\begin{table}[!htbp]
\centering
\fontsize{8}{11}\selectfont
\setlength{\tabcolsep}{2pt}
\begin{tabular}{@{}lccccc@{}}
\toprule
\textbf{Model} & \textbf{MMLU} & \textbf{split-half} & \textbf{cat 1-NN} & \textbf{any-hit@3} & \textbf{med.\ VE} \\
\midrule
Qwen2.5-0.5B & 47.5 & 0.95 & 0.67 & 0.67 & 0.990 \\
Qwen2.5-1.5B & 60.9 & 0.97 & 0.68 & 0.75 & 0.997 \\
Qwen2.5-3B   & 65.6 & 0.97 & 0.72 & 0.75 & 0.992 \\
Qwen2.5-7B   & \textbf{74.2} & 1.00 & \textbf{0.74} & \textbf{0.78} & 0.998 \\
\midrule
Llama-3.2-1B & 32.2 & 0.93 & 0.63 & 0.68 & 0.993 \\
Llama-3.2-3B & 58.0 & 1.00 & 0.74 & 0.73 & 0.990 \\
Llama-3.1-8B & \textbf{66.7} & 1.00 & \textbf{0.74} & \textbf{0.75} & 0.987 \\
\bottomrule
\end{tabular}
\caption{Per-model fingerprint metrics at $K{=}50$, grouped by family and ordered by size within each family (MMLU shown as context). split-half: fingerprint self-retrieval top-1 (chance $0.02$); cat 1-NN: category nearest-neighbour purity (chance $0.24$); any-hit@3: fraction of subjects with $\geq 1$ of their top-3 dynamics neighbours among their top-3 semantic neighbours (chance $0.16$); med.\ VE: median per-step variance explained.}
\label{tab:models}
\end{table}

\textbf{Fingerprint quality is consistent across the seven models tested.} Every model is highly reproducible (split-half $0.93$-$1.00$ vs.\ $0.02$ chance), category-coherent (1-NN purity $0.63$-$0.74$ vs.\ $0.24$), and semantically aligned (any-hit@3 $0.67$-$0.78$ vs.\ $0.16$), with a near-linear interior throughout (median per-step variance explained $0.987$-$0.998$). These ranges span both families and more than an order of magnitude in scale (0.5-8B). We do not test whether two models assign similar mode subspaces to the same topic; the consistency is in the quality of each model's fingerprints. The evidence also covers only dense, decoder-only Qwen and Llama models, and the two families differ in detail, for example in the shape of the variance-explained profile near the output (Figure~\ref{fig:near-linearity}).

\paragraph{Fingerprint quality and capability.} We also compare the fingerprint metrics with the models' measured capability. Within each family both metrics rise monotonically with size (cat 1-NN $0.67\!\to\!0.74$ and any-hit@3 $0.67\!\to\!0.78$ for Qwen 0.5-7B; $0.63\!\to\!0.74$ and $0.68\!\to\!0.75$ for Llama 1-8B), and across all seven models they track published 5-shot MMLU: the rank correlation with MMLU is Spearman $\rho \approx 0.79$ for cat 1-NN and $\rho \approx 0.93$ for any-hit@3. With only seven models this is suggestive rather than conclusive, and the correlations rest heavily on the lowest-MMLU model (Llama-3.2-1B); we therefore treat the link between fingerprint quality and capability as a hypothesis for larger model sets.

\section{Conclusion}

We fitted a linear Koopman operator to the layer-to-layer evolution of hidden states in seven transformer language models and used its dominant mode subspace as a per-topic spectral fingerprint. The depth map was found to be \emph{near-linear} in its interior, with its nonlinearity concentrated at the input and output layers. The resulting fingerprint is \emph{topic-discriminative}: aligned with both the MMLU taxonomy and semantic neighbourhoods, while grouping quantitative subjects across categories. The spectral fingerprint can, therefore, be seen as a probe of the structure beneath a model's behaviour, and a starting point for future interventions. A natural extension is to fit separate operators to blocks of layers, which could separate the roles of different stages of the network, such as the nonlinear input and output stages and the near-linear interior we observe.

\section*{Limitations}

\paragraph{Untested model classes.} Our cross-model evidence covers seven dense, decoder-only base models from two families, up to 8B parameters. We leave models with substantially different attention or normalisation designs, mixture-of-experts models, instruction-tuned or RLHF-trained variants, and larger models as future work.

\paragraph{Design choices and causality.} The operator rank (100) and the fingerprint dimension ($K{=}50$) are empirical choices; $K$ was selected from a sweep on one model and then applied to all, rather than tuned per model. Our evidence is correlational throughout: we show that fingerprint geometry aligns with taxonomic and semantic structure, but we do not intervene on the identified directions, so we cannot establish that the model causally relies on them. All experiments use English; whether the results transfer to other languages is an open question.

\paragraph{Static baselines.} We do not compare the fingerprint against static topic representations that ignore the layer-to-layer pairing, such as the mean hidden state of a topic or the principal subspace of its hidden states. Such representations may also recover the taxonomic and semantic alignment we report. The dynamic analysis is, however, required for the one-step near-linearity and interior-boundary findings (Section~\ref{sec:recon57}), which static representations cannot express.

\bibliography{custom}

\end{document}